\documentclass{article}
\usepackage{ijcai26}

\usepackage{times}
\usepackage{soul}
\usepackage{url}
\usepackage[hidelinks]{hyperref}
\usepackage[utf8]{inputenc}
\usepackage[small]{caption}
\usepackage{graphicx}
\usepackage{amsmath}
\usepackage{amsthm}
\usepackage{booktabs}
\usepackage{algorithm}
\usepackage{algorithmic}
\usepackage[switch]{lineno}

\usepackage{multirow}
\usepackage[table,xcdraw]{xcolor}
\usepackage{tcolorbox}
\usepackage[utf8]{inputenc} % allow utf-8 input
\usepackage[T1]{fontenc}    % use 8-bit T1 fonts
\usepackage{url}            % simple URL typesetting
\usepackage{booktabs}       % professional-quality tables
\usepackage{pifont}
\usepackage{amsfonts}       % blackboard math symbols
\usepackage{nicefrac}       % compact symbols for 1/2, etc.
\usepackage{microtype}      % microtypography
\usepackage{enumitem}
\usepackage{adjustbox}

\DeclareMathOperator*{\argmin}{arg\,min}

\usepackage{newfloat}
\usepackage{listings}
\DeclareCaptionStyle{ruled}{labelfont=normalfont,labelsep=colon,strut=off} % DO NOT CHANGE THIS
\floatstyle{ruled}
\newfloat{listing}{tb}{lst}{}
\floatname{listing}{Listing}

\newtheorem{theorem}{Theorem}

\title{\protect{\includegraphics[width=1.0cm]{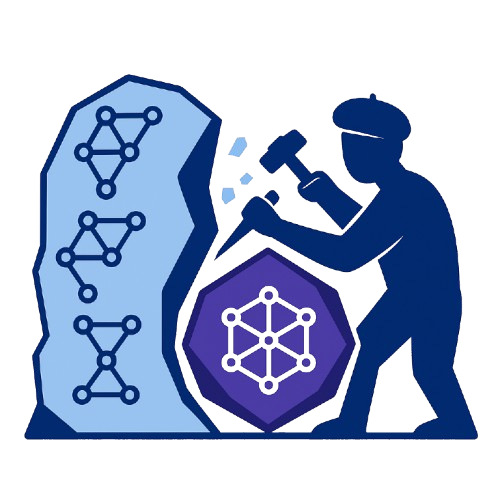}} GraphSculptor: Sculpting Pre-training Coreset \\ for Graph Self-supervised Learning}

\author{
Chuang Liu$^1$
\and
Zelin Yao$^2$\and
Xueqi Ma$^3$\and
Luzhi Wang$^4$\and
Mukun Chen$^5$\\
Pinghua Xu$^1$\And
Wenbin Hu$^2$\\
\affiliations
$^1$Sangfor Technologies Inc.\\
$^2$Wuhan University\\ 
$^3$The University of Melbourne\\
$^4$Dalian Maritime University\\
$^5$Hunan University of Technology and Business; Xiangjiang Laboratory\\
\emails
\{chuangliu, zelinyao, cmk0910, xupinghua, hwb\}@whu.edu.cn\\
xueqim@student.unimelb.edu.au, wangluzhi0@gmail.com
}
\begin{document}

\maketitle

\begin{abstract}
Graph self-supervised learning (SSL) typically relies on large-scale unlabeled datasets, heavily inflating computational costs. However, empirical evidence suggests that these datasets contain substantial redundancy—our analysis reveals that uniformly subsampling 50\% of graphs retains over 96\% of downstream performance. To exploit this redundancy, we introduce GraphSculptor for pre-training coreset construction. Unlike methods dependent on additional training-time signals or limited solely to topological statistics, GraphSculptor provides a label-free solution that constructs coresets via two complementary perspectives: intrinsic structure and contextual semantics. Concretely, structural diversity is quantified using intrinsic graph statistics, yielding a structural feature vector for each graph, while semantic diversity is captured by utilizing a pre-trained language model to encode descriptions generated via graph-to-text. GraphSculptor integrates these signals into a unified metric space and performs cluster-aware selection to preserve joint structural--semantic diversity. We further derive a theoretical bound on the loss gap between coreset and full-data pre-training, offering theoretical motivation for our selection formulation. Extensive experiments demonstrate that GraphSculptor effectively ``sculpts'' the dataset: a 10\% coreset achieves 99.6\% of full-data performance while reducing pre-training time by nearly 90\%, offering a scalable solution for data-efficient graph pre-training. 
\end{abstract}

\section{Introduction}
\label{sec:introduction}
Graph self-supervised learning (SSL) has become a cornerstone for representation learning on graph-structured data, achieving strong performance across domains from molecular science to social network analysis~\cite{graphcl,simgrace,graphmae}. Recent progress is often driven by scaling up unlabeled pre-training datasets, implicitly assuming that more pre-training graphs will reliably translate into better representations. In this work, we revisit this assumption and show that large graph datasets can contain substantial redundancy, motivating a more compute-efficient, data-centric alternative.

\begin{figure}[!t] % !ht% \setlength{\abovecaptionskip}{-0.1cm}   %调整图片标题与图距离
\begin{center}
\includegraphics[width=1.0\linewidth]{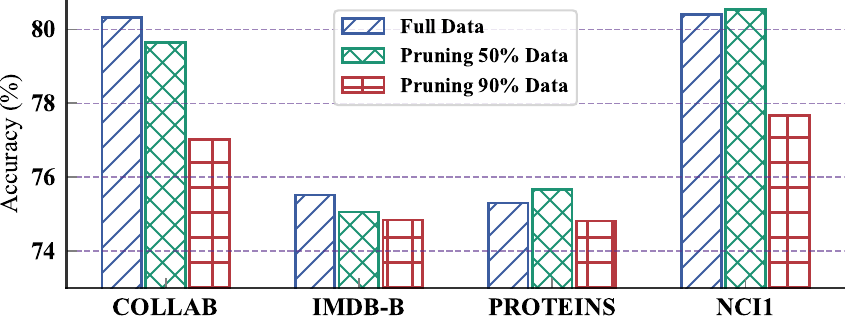}
\end{center}
\caption{\textbf{Substantial redundancy in graph SSL datasets.}
We pre-train GraphMAE on uniformly subsampled subsets (retaining 50\% and 10\% of graphs) and evaluate on four benchmark datasets. Performance degrades only mildly under random pruning; in particular, when retaining 50\% of the pre-training graphs, the worst-case relative drop across these datasets is below 4\%. This observation suggests that a sizeable fraction of the dataset provides marginal utility for pre-training and motivates principled coreset curation.}
\label{fig:motiv}
\end{figure}

We begin with a simple diagnostic experiment (Figure~\ref{fig:motiv}) showing that graph SSL pre-training can be empirically robust to uniform random pruning in our controlled setup. Specifically, randomly removing 50\% of the pre-training graphs preserves over 96\% of the downstream transfer performance. This result suggests that, at least in this setting, scaling pre-training data may exhibit diminishing returns and that a substantial portion of graphs can be redundant for learning transferable representations. Such redundancy not only wastes computation, but may also skew pre-training toward frequent and low-variation patterns. This naturally raises the following question: \emph{can we identify a much smaller subset that remains representative of the full pre-training dataset?}

To answer this question, we take a data-centric view of graph SSL: we aim to improve pre-training efficiency by selecting informative pre-training graphs instead of increasing data scale. Prior efforts to reduce training data often rely on training-time signals such as gradients or task supervision to identify ``important'' examples~\cite{apt,gder}. 
However, in graph SSL pre-training, labels are typically unavailable, and computing gradient-based criteria can introduce additional optimization overhead and may be tied to specific model/optimizer choices, potentially limiting the achievable wall-clock savings and generality. To this end, we introduce \textbf{GraphSculptor}, a label-free and model-agnostic framework that extracts an information-rich coreset from a large pre-training collection. GraphSculptor measures diversity from two complementary views---structure via intrinsic graph statistics and semantics via graph-to-text followed by a frozen text encoder---and then selects a coreset that preserves joint structural--semantic diversity under a target budget. Our objective is to substantially reduce pre-training cost while maintaining downstream transfer performance comparable to full-data pre-training.

\begin{figure}[!t] % !htb
\begin{center}
\includegraphics[width=0.95\linewidth]{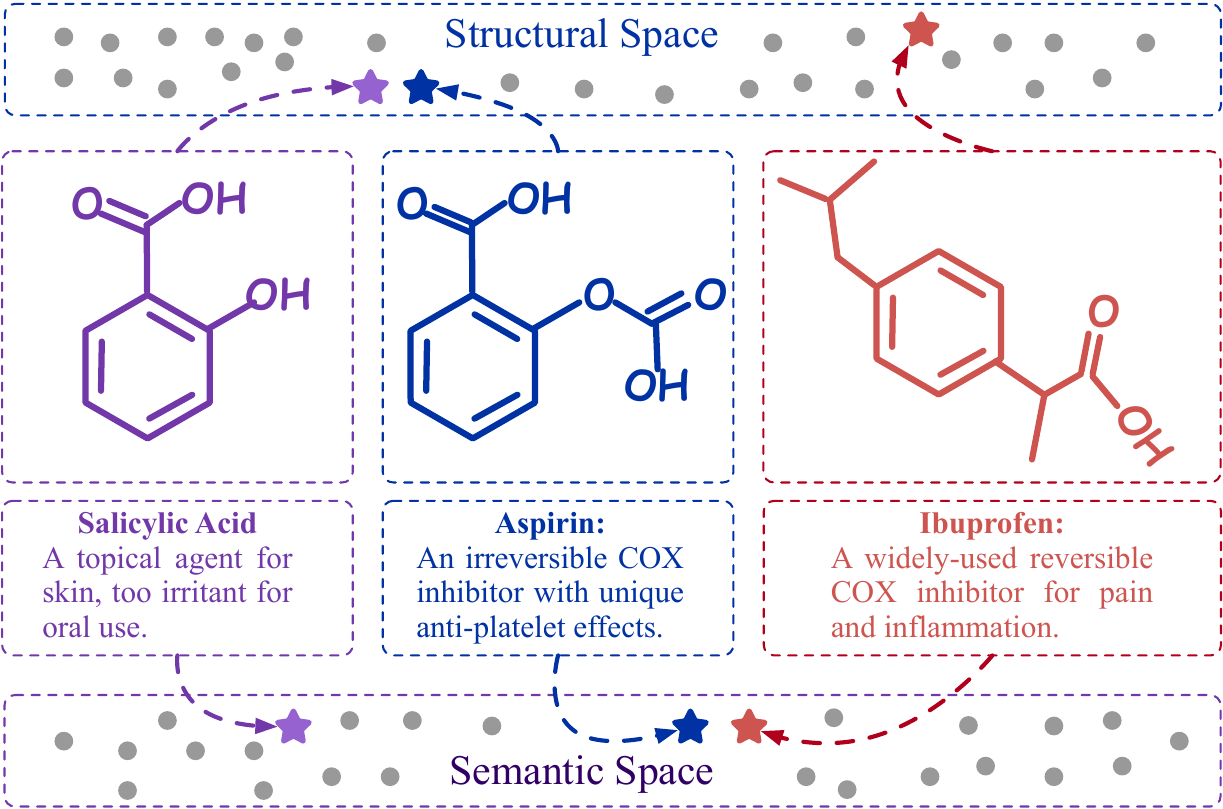}
\end{center}
\caption{\textbf{Structure vs.\ semantics can disagree.}
Molecules that are structurally similar (\textit{e.g.}, Aspirin vs.\ Salicylic Acid) may be semantically distant in terms of their textual descriptions and associated functions, while structurally dissimilar ones (\textit{e.g.}, Aspirin vs.\ Ibuprofen) can be semantically closer. This motivates measuring diversity beyond topology by indatasetsting semantic embeddings derived from graph-to-text.}
\label{fig:semantic}
\end{figure}

The core idea of GraphSculptor is to curate pre-training data using a dual-view notion of diversity. We characterize each graph from two complementary perspectives: \textbf{structure} and \textbf{semantics}. For the structural view, we compute a compact vector of intrinsic, multi-scale graph statistics and aim to cover diverse topological regimes, rather than over-representing the most frequent patterns. At the same time, structure alone may be insufficient: in many domains, graphs with similar topology can correspond to different domain properties (\textit{e.g.}, molecules sharing similar scaffolds may have different biological activities; see Figure~\ref{fig:semantic}). To capture such domain-level variation, we map each graph to a textual description (graph-to-text) and encode it with a pre-trained language model, yielding a semantic embedding that serves as a complementary proxy beyond topology-based summaries.

Given the fused structural and semantic embeddings, GraphSculptor selects a budgeted coreset in a cluster-aware manner. Specifically, it clusters graphs in the joint space and then selects representative graphs across clusters, prioritizing clusters that are both diverse internally and well separated from others, to preserve joint structural--semantic diversity under the target budget. The selection in GraphSculptor is performed once per pre-training dataset as an offline preprocessing step: we compute structural/semantic embeddings for all graphs and derive a fixed coreset that can be reused across pre-training runs. This one-time cost is amortized when the same dataset is used to pre-train different models or under different SSL objectives, and it avoids any training-time gradient-based scoring, making the procedure label-free, model-agnostic, and easy to cache and deploy. Our main contributions are:

\begin{itemize}[leftmargin=1em]
    \item We propose GraphSculptor, a label-free and model-agnostic coreset construction framework for graph SSL that jointly considers structural statistics and semantic embeddings.
    \item We provide a loss-gap bound for coreset pre-training versus full-data pre-training, supporting our selection principle.
    \item Experiments show strong efficiency--accuracy trade-offs: using 10\% of the pre-training graphs retains up to 99.6\% of full-data performance while reducing pre-training time by nearly 90\%.
\end{itemize}

\section{Related Work}
\label{sec:preliminary}

\paragraph{Graph Pretraining.}
Self-supervised graph pre-training has become a standard approach for leveraging unlabeled graphs and mitigating label scarcity across domains~\cite{survey-pretraining}. Existing methods broadly fall into two paradigms: contrastive pre-training, which aligns representations across augmented views to learn invariances~\cite{graphcl,simgrace}, and generative pre-training, which learns by reconstructing masked components of graphs~\cite{maskgae,hi-gmae,structure-mae}. The latter line, represented by GraphMAE and its variants~\cite{graphmae,graphmae2}, is often adopted due to its relatively simple training pipeline without explicit negative sampling. Subsequent work extends the reconstruction targets to richer structural signals (\textit{e.g.}, edges and degrees)~\cite{maskgae,s2gae,gigamae,graphpae} or combines reconstruction with contrastive objectives~\cite{simsgt,protomgae,aug-mae,acgmae}. While these approaches differ in objectives and architectures, they typically pre-train on the full available dataset and benefit from larger-scale data when feasible. This practice can incur substantial computational cost and limits the iteration speed and scalability of graph pre-training. In contrast, our work takes a data-centric perspective and asks how to curate a compact yet informative pre-training subset that preserves the utility of full-dataset pre-training. Therefore, our method is orthogonal to the choice of pre-training objective and can be applied on top of existing graph pre-training frameworks.

\paragraph{Graph Data Reduction.}
The high computational cost of learning on large-scale graph datasets has motivated research on reducing training data, broadly including graph dataset distillation/condensation and graph selection. Distillation methods such as GCond~\cite{gcond} and DosCond~\cite{doscond} aim to synthesize a compact surrogate dataset, often formulated as bilevel optimization to match training dynamics (\textit{e.g.}, gradients). While effective in some settings, these formulations can be computationally demanding and sensitive to modeling choices, prompting more tractable alternatives: KIDD~\cite{kidd} replaces gradient matching with kernel-based objectives, SGDD~\cite{sgdd} emphasizes preserving structural properties, and MIRAGE~\cite{mirage} condenses representative substructures with improved model-agnosticism. Graph selection, which directly chooses a subset from the original dataset, is closer to our setting. Many existing selection approaches adopt dynamic in-training strategies that iteratively update the subset using model-dependent signals, such as training loss~\cite{molpeg}, gradient norms~\cite{gder}, or predictive uncertainty~\cite{apt}. Although these signals can be informative, the resulting procedures are tightly coupled to specific training dynamics and typically require repeated evaluations over training, increasing overhead and reducing their suitability as a one-shot, model-agnostic preprocessing step~\cite{gradate}. In contrast, GraphSculptor performs one-shot, offline coreset selection without training-time signals, enabling label-free and model-agnostic preprocessing before pre-training.

\begin{figure*}[!t] % !htb
\begin{center}
\includegraphics[width=0.95\linewidth]{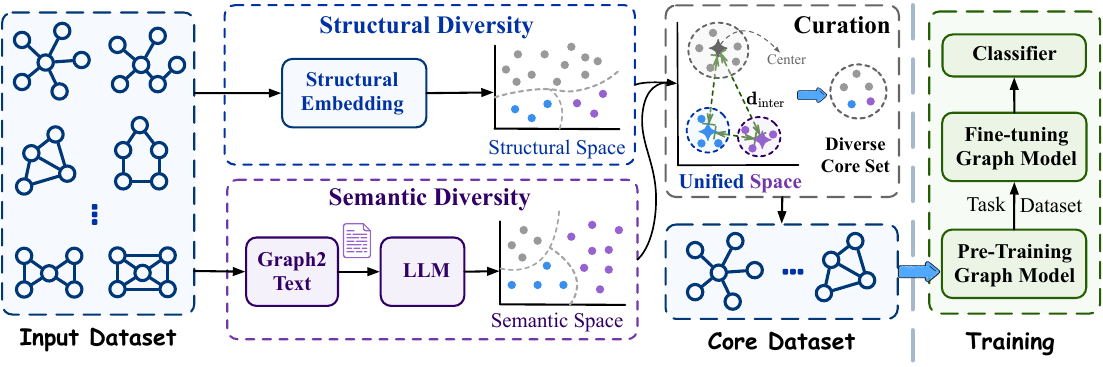}
\end{center}
\caption{\textbf{The GraphSculptor workflow.} We construct a coreset by jointly optimizing for Structural Diversity and  Semantic Diversity. These are fused into a unified space where our curation module selects an information-rich subset for efficient and effective pre-training.}
\label{fig:model}
\end{figure*}

\section{Method}
\label{sec:method}

GraphSculptor is a data-centric framework for constructing a compact pre-training coreset from a large unlabeled graph dataset. As illustrated in Figure~\ref{fig:model}, it follows two stages: (1) \textbf{Holistic Graph Representation}, which computes for each graph a fused representation that combines structural and semantic views; and (2) \textbf{Principled Coreset Curation}, which selects a budgeted subset via cluster-based selection in the fused space to preserve structural--semantic diversity. In particular, the semantic view is obtained by mapping each graph to a textual description (graph-to-text) and encoding it with a frozen text encoder, providing signals complementary to topology-based statistics. The entire procedure is performed offline as a one-time preprocessing step before SSL pre-training.

\paragraph{Preliminaries.}
We consider an unlabeled graph dataset $\mathcal{D}=\{G_i\}_{i=1}^{M}$, where each graph $G_i=(\mathcal{V}_i,\mathcal{E}_i)$ is associated with an adjacency matrix $\mathbf{A}_i$ and node features $\mathbf{X}_i$. Our goal is to select a coreset $\mathcal{D}_{\text{core}}\subset\mathcal{D}$ with a fixed budget
$|\mathcal{D}_{\text{core}}| = M_{\text{target}}=\lfloor p_{\text{target}}\,M\rfloor$,
where $p_{\text{target}}\in(0,1]$ denotes the retention ratio. We construct $\mathcal{D}_{\text{core}}$ to remain representative of $\mathcal{D}$ in the proposed fused (structural--semantic) representation space under this budget, \textit{i.e.}, to preserve the diversity of graphs as measured in that space.

\subsection{Holistic Graph Representation}
\label{sec:representation}

We represent each graph by jointly capturing (i) intrinsic topology and (ii) semantic context. Concretely, for each $G_i$ we compute a structural feature vector $\mathbf{z}_{S,i}\in\mathbb{R}^{d_S}$ and a semantic embedding $\mathbf{z}_{T,i}\in\mathbb{R}^{d_T}$, where $d_S$ and $d_T$ denote their respective feature dimensions. We then combine them into a fused representation used for downstream clustering and coreset selection.

\subsubsection{Structural Characterization}
\label{sec:topo_char}

To capture structural diversity, we characterize each graph using a compact set of graph-level descriptors. The design goal is to balance scalability and discriminative power: the descriptors should be inexpensive to compute at dataset scale while still reflecting salient structural variations relevant to data curation. We therefore adopt a modular structural feature extractor that can incorporate domain-specific descriptors when beneficial, while using a lightweight default configuration in our main experiments.

For each graph $G_i$, we define a structural feature map $\psi_S:\mathcal{G}\rightarrow\mathbb{R}^{d_S}$ that outputs a fixed-dimensional descriptor $\mathbf{z}_{S,i}$. Specifically,
\begin{equation}
\label{eq:struct_feat_generalized}
\mathbf{z}_{S,i}=\psi_S(G_i)=\left[\phi_1(G_i),\phi_2(G_i),\dots,\phi_{d_S}(G_i)\right]^\top,
\end{equation}
where each $\phi_j:\mathcal{G}\rightarrow\mathbb{R}$ computes a scalar structural statistic. Concretely, $\psi_S$ is instantiated with a streamlined set of descriptors, engineered to strike an optimal balance between discriminative expressivity and computational scalability:
\noindent (i) Basic Topological Invariants. We explicitly encode the fundamental scale and sparsity of each graph via its node count $|\mathcal{V}_i|$, edge count $|\mathcal{E}_i|$, and average degree $2|\mathcal{E}_i|/|\mathcal{V}_i|$.
\noindent (ii) Global Positional Encodings. To transcend local aggregation metrics and capture long-range structural dependencies, we incorporate Positional Encodings. Specifically, we leverage random-walk based positional identifiers~\cite{graphormer-v1}, which provide a compact yet informative signature of global connectivity, ensuring the structural feature vector remains distinct across topologically diverse graphs.

The resulting $\mathbf{z}_{S,i}$ serves as the structural view used in the fused representation. We use a lightweight default configuration in our main experiments for efficiency, and $\psi_S$ can be extended with additional domain-specific descriptors when they are available and useful.

\subsubsection{Semantic Contextualization}
\label{sec:sem_context}

Beyond structure, many graph datasets carry domain semantics that may not be fully captured by structural descriptors alone. For molecular graphs, compounds with highly
similar scaffolds can exhibit markedly different biochemical effects (\textit{e.g.}, Aspirin vs.\ Salicylic Acid). Similarly, in citation networks, two papers may have nearly
identical local citation neighborhoods (\textit{e.g.}, comparable in-/out-degree and co-citation patterns) while belonging to entirely different research topics. To
indatasetste this complementary signal, we leverage natural-language descriptions by mapping each graph to text and then encoding the text with a pre-trained text encoder,
yielding a semantic embedding that is orthogonal to purely structural statistics. We define a semantic feature map as a two-stage composition:
\begin{equation}
  \psi_T \;=\; f_{\text{T2E}} \circ f_{\text{G2T}},
\end{equation}
which maps each graph to a dense semantic embedding.

\paragraph{Graph-to-Text conversion ($f_{\text{G2T}}$).} We map each graph $G_i$ to a short textual descriptor $t_i=f_{\text{G2T}}(G_i)$ using domain-specific instantiations. For molecular graphs, we leverage the available SMILES strings and employ a biomedical text generation model (\textit{e.g.}, BioT5~\cite{biot5}) to generate a concise title/description. For social-network graphs, where no canonical textual field is provided, we first serialize the graph into an adjacency-list (edge-list) representation with a fixed, deterministic ordering (\textit{e.g.}, sorting nodes and neighbors by index to reduce permutation effects). We then prepend a fixed dataset-level background description that specifies only the domain context (without labels or task information), and use a general-purpose LLM (\textit{e.g.}, ChatGPT or Gemini) to produce a bounded-length, sociology-oriented title. We use a fixed prompt template and deterministic decoding, and cache the generated texts to ensure reproducibility.

\paragraph{Text-to-Embedding encoding ($f_{\text{T2E}}$).} We encode $t_i$ with a pre-trained text encoder to obtain a semantic embedding.
In our main experiments on biochemical graphs, we use SciBERT~\cite{scibert} as the encoder; for general text, standard encoders such as BERT~\cite{bert} can be used.
Formally, given a text encoder $\text{Encoder}(\cdot)$, we use the final-layer hidden state of the special classification token \texttt{[CLS]} as a pooled representation:
\begin{equation}
\label{eq:semantic_embed}
\mathbf{z}_{T,i}=f_{\text{T2E}}(t_i)=\text{Encoder}(t_i)_{[\text{CLS}]}
\in \mathbb{R}^{d_T}.
\end{equation}
During coreset construction, the text encoder is kept frozen to avoid introducing any training-time signals and to keep the procedure lightweight. The resulting embeddings provide a semantics-aware proxy derived from natural language, complementing structural descriptors by capturing descriptive similarity that may not be reflected by topology alone.

\subsubsection{Unified Representation Space}
\label{sec:fusion}

To enable holistic coreset selection, we project the distinct structural and semantic signals into a unified metric space. Direct fusion of the structural descriptor $\mathbf{z}_{S,i} \in \mathbb{R}^{d_S}$ and the semantic embedding $\mathbf{z}_{T,i} \in \mathbb{R}^{d_T}$ is non-trivial due to their disparate modalities and potentially incommensurable magnitudes. To prevent high-variance features from dominating the distance computation, we employ feature-wise standardization (z-score normalization) across the dataset, yielding normalized descriptors $\hat{\mathbf{z}}_{S,i}$ and $\hat{\mathbf{z}}_{T,i}$. We then synthesize the holistic representation $\mathbf{h}_i$ via concatenation:
\begin{equation}
\mathbf{h}_i = \text{Concat}\!\left(\hat{\mathbf{z}}_{S,i}, \, \hat{\mathbf{z}}_{T,i}\right).
\end{equation}
In this unified space, the Euclidean distance aggregates both structural similarity and semantic affinity, preventing modal dominance and ensuring that coreset selection targets joint structural--semantic diversity.

\subsection{Principled Coreset Curation}
\label{sec:curation}

% \subsubsection{Diversity-Aware Selection}
% \label{sec:sampling}

We construct the coreset via a cluster-aware selection scheme in the fused space. We first partition graphs by $K$-means, then prioritize clusters using a diversity score, and finally select representative graphs within each cluster under a fixed budget.

\paragraph{\text{\large \ding{202}} Unsupervised Clustering in the Fused Space.}
Given fused representations $\{\mathbf{h}_i\}_{i=1}^{M}$, we apply $K$-means clustering and obtain clusters $\mathcal{C}=\{C_1,\dots,C_K\}$ with centroids $\{\boldsymbol{\mu}_k\}_{k=1}^{K}$ by minimizing the within-cluster squared Euclidean distance:
\begin{equation}
\label{eq:kmeans_holistic}
\underset{\mathcal{C}}{\argmin}\ \sum_{k=1}^{K}\sum_{\mathbf{h}_i\in C_k}\left\|\mathbf{h}_i-\boldsymbol{\mu}_k\right\|_2^2.
\end{equation}

\paragraph{\text{\large \ding{203}} Diversity Scoring.}
An effective coreset should balance \emph{mass coverage} (so that high-density regions of the empirical distribution are not under-represented) and \emph{diversity} to avoid collapsing onto frequent but low-informative patterns. To combine these factors in a simple and scale-robust manner, we define the cluster score in the log domain:
\begin{equation}
\label{eq:complexity_score_adv}
\Omega_j=\log(\pi_j)+w\log(d_{\text{intra},j})+(1-w)\log(d_{\text{inter},j}),
\end{equation}
where $\pi_j=|C_j|/M$ captures the cluster mass and $w\in[0,1]$ controls the trade-off between within-cluster dispersion and cross-cluster separation. This log-domain formulation ensures numerical stability and balances the magnitude differences between density and structural metrics. We next normalize these weights to obtain cluster-wise sampling proportions.

\paragraph{\text{\large \ding{204}} Cluster-wise Selection.}
We convert $\{\Omega_j\}$ into cluster-wise sampling proportions:
\begin{equation}
\label{eq:sampling_prob}
\tilde{\Omega}_j=\frac{\exp(\Omega_j/\tau)}{\sum_{k=1}^{K}\exp(\Omega_k/\tau)},\qquad \tau>0,
\end{equation}
where we use $\tau=0.5$ by default. We then assign each cluster a quota $n_j$ such that $\sum_{j=1}^{K}n_j=M_{\text{target}}$; in implementation, we set $n_j=\lfloor M_{\text{target}}\tilde{\Omega}_j\rfloor$ for $j=1,\dots,K-1$ and allocate the remainder to the last cluster.

Within each cluster $C_j$, we perform centroid-biased sampling to obtain $\mathcal{S}_j$.
For $G_i\in C_j$, we define
\begin{equation}\label{eq:soft_proto_prob}
p_{i\mid j}=\frac{\exp\!\left(-\frac{\|\mathbf{h}_i-\boldsymbol{\mu}_j\|_2^2}{2\sigma_j^2}\right)}
{\sum_{i'\in C_j}\exp\!\left(-\frac{\|\mathbf{h}_{i'}-\boldsymbol{\mu}_j\|_2^2}{2\sigma_j^2}\right)}.
\end{equation}
and sample $n_j$ graphs without replacement from $C_j$ according to $\{p_{i\mid j}\}$:
\begin{equation}\label{eq:core_final}
\mathcal{D}_{\text{core}}=\bigcup_{j=1}^{K}\mathcal{S}_j,\qquad
\mathcal{S}_j\sim \textsc{WRS}\!\left(C_j, n_j; \{p_{i\mid j}\}\right),
\end{equation}
where $\textsc{WRS}(\cdot)$ denotes weighted random sampling without replacement, \textit{i.e.}, drawing $n_j$ samples from $C_j$ with probabilities proportional to $\{p_{i\mid j}\}$.
Since global coverage is mainly ensured by the cross-cluster budget allocation (Eq.~(\ref{eq:complexity_score_adv})), this within-cluster step favors typical instances near the centroid while retaining randomness. The temperature $\sigma_j$ controls the sampling sharpness, interpolating between uniform sampling ($\sigma_j\!\to\!\infty$) and hard nearest-to-centroid selection ($\sigma_j\!\to\!0$).

\subsection{Theoretical Analysis}
We provide theoretical motivation for our cluster-aware allocation by analyzing optimization under a standard NTK-style linearization around the initialization $W_0$. Here, $W$ denotes the parameters of the pre-training model and $W_0$ is their initialization. Concretely, we consider the linearized per-sample loss
$\tilde{\ell}(W;G)=\ell(W_0;G)+\langle \nabla \ell(W_0;G),\, W-W_0\rangle$, so that the optimization dynamics are governed by initialization gradients. Under $L$-smoothness and bounded-gradient assumptions, we bound the expected pre-training loss gap between training on the selected coreset and training on the full dataset. The bound highlights the role of within-cluster gradient variability (heterogeneity), suggesting that clusters with larger variability should receive higher sampling proportions.

\begin{theorem}[\textbf{Expected Pre-training Loss Gap under Linearization}]
\label{thm:main_result}
Let $\mathcal{L}(W)=\frac{1}{M}\sum_{i=1}^{M}\ell(W;G_i)$ be an $L$-smooth loss over $\mathcal{D}=\{G_1,\dots,G_M\}$. Assume $\|\nabla \ell(W_0;G)\|_F\le B_g$ for all $G\in\mathcal{D}$ and the linearized regime defined by $\tilde{\ell}(\cdot;\cdot)$ above. Suppose $\mathcal{D}$ is partitioned into $K$ disjoint clusters $\{C_k\}_{k=1}^K$ with proportions $\pi_k=|C_k|/M$. Let $M_{\text{target}}$ be the coreset size, and let $M_k$ be the number of selected samples from cluster $C_k$, inducing $q_k=M_k/M_{\text{target}}$ (with $q_k>0$ and $\sum_k q_k=1$). Define the within-cluster gradient variance at $W_0$ as
\[
\mathcal{V}_k^2=
\mathbb{E}_{G\in C_k}\!\left[
\left\|\nabla \ell(W_0;G)-\mathbb{E}_{G'\in C_k}[\nabla \ell(W_0;G')]\right\|_F^2
\right].
\]
Let $W^{(T)}_{\text{full}}$ and $W^{(T)}_{\text{core}}$ denote the parameters after $T$ gradient steps with learning rate $\eta$ when optimizing on the full dataset and on the sampled coreset, respectively. Then the expected pre-training loss gap satisfies:
\begin{equation}
\label{eq:theorem}
\mathbb{E}_{\mathcal{D}_{\text{core}}}\!\left[
\mathcal{L}(W^{(T)}_{\text{core}})-\mathcal{L}(W^{(T)}_{\text{full}})
\right]
\le
\frac{L T^2 \eta^2}{2\,M_{\text{target}}}
\sum_{k=1}^{K}\frac{\pi_k^2\,\mathcal{V}_k^2}{q_k}.
\end{equation}
\end{theorem}

\noindent\textbf{Remark.} Minimizing the upper bound in Eq.~(\ref{eq:theorem}) subject to $\sum_{k=1}^{K} q_k=1$ yields $q_k^\star \propto \pi_k \mathcal{V}_k$, \textit{i.e.}, larger clusters with higher gradient variability should receive more budget.

\paragraph{Discussion.}
In practice, $\mathcal{V}_k$ is not available without training-time gradient evaluations. We therefore adopt offline, representation-based proxies computed once in our fused structural--semantic space. Under a mild regularity condition that initialization gradients vary smoothly with the fused representation (\textit{e.g.}, $\|\nabla \ell(W_0;G_i)-\nabla \ell(W_0;G_{i'})\|_F \le L_h \|\mathbf{h}_i-\mathbf{h}_{i'}\|_2$), the within-cluster gradient variability can be controlled by within-cluster dispersion in the fused space, motivating the mass--intra component $\log(\pi_k)+w\log(d_{\text{intra},k})$. We further incorporate inter-cluster separation $d_{\text{inter},k}$ as a practical correction to discourage allocating excessive budget to highly overlapping clusters, leading to our score
$\Omega_k$
(Eq.~(\ref{eq:complexity_score_adv})). Sampling proportions are then obtained via temperature-scaled softmax, implementing the mass--variability principle suggested by the theory while remaining label-free and avoiding training-time signals.

\definecolor{lightpink}{HTML}{FAEAE1}
\definecolor{iceblue}{HTML}{E0F5FF}
\tcbset{
  pinkbox/.style={
    colback=lightpink,
    colframe=lightpink,
    width = 0.8cm,
    height = 0.35cm,
    halign=center, % 居中对齐
    valign=center, % 竖直居中
  }
}
\tcbset{
  bluebox/.style={
    colback=iceblue,
    colframe=iceblue,
    width = 0.8cm,
    height = 0.35cm,
    halign=center, % 居中对齐
    valign=center, % 竖直居中
  }
}

\begin{table*}[!t]
\centering
\caption{Transfer learning performance on 8 MoleculeNet benchmarks (ROC-AUC \%). Results are grouped by sampling ratio. Full Dataset refers to the performance using the original graph dataset. \textbf{Bold} indicates the best result within each sampling ratio.}
\label{tab:trans}
\renewcommand\arraystretch{1.02}
\setlength\tabcolsep{3pt}
\resizebox{0.99\textwidth}{!}{%
\begin{tabular}{@{}llccccccccc@{}}
\toprule
\textbf{Data Remaining} & \textbf{Method} & BBBP & Tox21 & ToxCast & SIDER & Clintox & MUV & HIV & BACE & \textbf{Avg.} \\
\midrule
 & Full Dataset
 & $71.7_{\pm 0.7}$ & $75.6_{\pm 0.6}$ & $63.8_{\pm 0.4}$ & $60.4_{\pm 1.0}$ & $81.0_{\pm 1.9}$ & $77.3_{\pm 2.1}$ & $76.5_{\pm 1.3}$ & $83.3_{\pm 0.8}$ & 73.7 \\
\midrule

\multirow{5}{*}{\shortstack{\textbf{70\% Dataset}\\\textbf{Remaining}}}
 & Random         & $70.8_{\pm 0.7}$ & $75.4_{\pm 0.1}$ & $64.1_{\pm 0.5}$ & $60.5_{\pm 0.7}$ & $81.8_{\pm 1.3}$ & $76.2_{\pm 0.9}$ & $77.1_{\pm 1.3}$ & $82.5_{\pm 0.6}$ & 73.6$_{\tiny\begin{tcolorbox}[bluebox, left=0pt]{\text{$\downarrow \textbf{0.1}$}} \end{tcolorbox}}$ \\
 & CD             & $71.5_{\pm 0.4}$ & $75.2_{\pm 0.4}$ & $64.0_{\pm 0.4}$ & $60.0_{\pm 0.7}$ & $84.2_{\pm 1.0}$ & $74.9_{\pm 1.9}$ & $76.7_{\pm 0.3}$ & $82.6_{\pm 1.3}$ & 73.6$_{\tiny\begin{tcolorbox}[bluebox, left=0pt]{\text{$\downarrow \textbf{0.1}$}} \end{tcolorbox}}$ \\
 & K-Center       & $72.5_{\pm 0.5}$ & $74.6_{\pm 0.6}$ & $63.5_{\pm 0.4}$ & $60.3_{\pm 0.5}$ & $82.9_{\pm 1.7}$ & $74.8_{\pm 0.9}$ & $77.1_{\pm 1.5}$ & $83.0_{\pm 1.2}$ & 73.6$_{\tiny\begin{tcolorbox}[bluebox, left=0pt]{\text{$\downarrow \textbf{0.1}$}} \end{tcolorbox}}$ \\
 & Herding        & $72.4_{\pm 0.4}$ & $75.3_{\pm 0.2}$ & $63.6_{\pm 0.3}$ & $59.0_{\pm 0.4}$ & $83.8_{\pm 1.0}$ & $75.1_{\pm 2.2}$ & $77.4_{\pm 0.6}$ & $82.7_{\pm 0.5}$ & 73.7$_{\tiny\begin{tcolorbox}[bluebox, left=0pt]{\text{$\downarrow \textbf{0.0}$}} \end{tcolorbox}}$ \\
 & \textbf{GraphSculptor}
 & $\textbf{72.8}_{\pm 0.4}$ & $75.3_{\pm 0.5}$ & $63.4_{\pm 0.4}$ & $\mathbf{62.5_{\pm 0.3}}$ & $81.9_{\pm 1.3}$ & $75.2_{\pm 1.6}$ & $\textbf{77.9}_{\pm 1.3}$ & $\textbf{83.8}_{\pm 1.0}$ & \textbf{74.1}$_{\tiny\begin{tcolorbox}[pinkbox, left=0pt]{\text{$\uparrow \textbf{0.4}$}} \end{tcolorbox}}$ \\
\midrule

\multirow{5}{*}{\shortstack{\textbf{50\% Dataset}\\\textbf{Remaining}}}
 & Random         & $71.4_{\pm 0.9}$ & $75.1_{\pm 0.5}$ & $63.5_{\pm 0.4}$ & $60.6_{\pm 0.7}$ & $83.4_{\pm 2.5}$ & $74.3_{\pm 2.0}$ & $77.1_{\pm 1.1}$ & $81.7_{\pm 0.5}$ & 73.4$_{\tiny\begin{tcolorbox}[bluebox, left=0pt]{\text{$\downarrow \textbf{0.3}$}} \end{tcolorbox}}$ \\
 & CD             & $71.8_{\pm 0.3}$ & $75.7_{\pm 0.7}$ & $63.3_{\pm 0.4}$ & $60.0_{\pm 0.7}$ & $83.7_{\pm 2.0}$ & $73.6_{\pm 1.8}$ & $77.2_{\pm 0.7}$ & $83.0_{\pm 1.5}$ & 73.4$_{\tiny\begin{tcolorbox}[bluebox, left=0pt]{\text{$\downarrow \textbf{0.3}$}} \end{tcolorbox}}$ \\
 & K-Center       & $72.4_{\pm 0.3}$ & $74.4_{\pm 0.4}$ & $63.5_{\pm 0.6}$ & $59.8_{\pm 1.1}$ & $83.2_{\pm 1.1}$ & $76.4_{\pm 1.9}$ & $76.2_{\pm 1.3}$ & $81.9_{\pm 0.7}$ & 73.5$_{\tiny\begin{tcolorbox}[bluebox, left=0pt]{\text{$\downarrow \textbf{0.2}$}} \end{tcolorbox}}$ \\
 & Herding        & $71.6_{\pm 1.0}$ & $74.5_{\pm 0.4}$ & $63.8_{\pm 0.3}$ & $60.6_{\pm 0.5}$ & $83.6_{\pm 1.4}$ & $74.8_{\pm 2.5}$ & $77.6_{\pm 0.5}$ & $82.1_{\pm 0.8}$ & 73.6$_{\tiny\begin{tcolorbox}[bluebox, left=0pt]{\text{$\downarrow \textbf{0.1}$}} \end{tcolorbox}}$ \\
 & \textbf{GraphSculptor}
 & $72.0_{\pm 0.4}$ & $\textbf{76.5}_{\pm 0.5}$ & $\textbf{64.3}_{\pm 0.5}$ & $60.5_{\pm 0.8}$ & $82.1_{\pm 1.5}$ & $74.4_{\pm 1.8}$ & $\textbf{77.9}_{\pm 1.3}$ & $82.8_{\pm 0.8}$ & \textbf{73.8}$_{\tiny\begin{tcolorbox}[pinkbox, left=0pt]{\text{$\uparrow \textbf{0.1}$}} \end{tcolorbox}}$ \\
\midrule

\multirow{5}{*}{\shortstack{\textbf{10\% Dataset}\\\textbf{Remaining}}}
 & Random         & $68.5_{\pm 0.8}$ & $74.2_{\pm 0.7}$ & $62.9_{\pm 0.7}$ & $59.4_{\pm 1.3}$ & $83.9_{\pm 1.4}$ & $73.1_{\pm 3.8}$ & $76.6_{\pm 1.5}$ & $80.5_{\pm 1.2}$ & 72.4$_{\tiny\begin{tcolorbox}[bluebox, left=0pt]{\text{$\downarrow \textbf{1.3}$}} \end{tcolorbox}}$ \\
 & CD             & $70.9_{\pm 0.9}$ & $74.0_{\pm 1.0}$ & $63.4_{\pm 0.8}$ & $59.0_{\pm 1.6}$ & $82.7_{\pm 2.1}$ & $73.6_{\pm 2.4}$ & $76.7_{\pm 1.7}$ & $80.5_{\pm 1.2}$ & 72.6$_{\tiny\begin{tcolorbox}[bluebox, left=0pt]{\text{$\downarrow \textbf{1.1}$}} \end{tcolorbox}}$ \\
 & K-Center       & $70.6_{\pm 0.9}$ & $74.2_{\pm 0.5}$ & $63.1_{\pm 0.4}$ & $59.7_{\pm 1.1}$ & $80.5_{\pm 3.0}$ & $73.8_{\pm 2.2}$ & $76.0_{\pm 1.0}$ & $81.0_{\pm 0.9}$ & 72.4$_{\tiny\begin{tcolorbox}[bluebox, left=0pt]{\text{$\downarrow \textbf{1.3}$}} \end{tcolorbox}}$ \\
 & Herding        & $70.5_{\pm 0.9}$ & $74.0_{\pm 0.6}$ & $63.1_{\pm 0.3}$ & $57.6_{\pm 1.6}$ & $83.5_{\pm 2.7}$ & $72.6_{\pm 1.4}$ & $76.7_{\pm 1.4}$ & $80.0_{\pm 1.1}$ & 72.3$_{\tiny\begin{tcolorbox}[bluebox, left=0pt]{\text{$\downarrow \textbf{1.4}$}} \end{tcolorbox}}$ \\
 & \textbf{GraphSculptor}
 & $\textbf{71.1}_{\pm 0.5}$ & $\textbf{74.7}_{\pm 0.7}$ & $\textbf{64.0}_{\pm 0.5}$ & $\textbf{60.1}_{\pm 1.1}$ & $\textbf{84.8}_{\pm 1.8}$ & $\textbf{74.0}_{\pm 1.2}$ & $\textbf{77.1}_{\pm 0.6}$ & $\textbf{81.2}_{\pm 0.6}$ & \textbf{73.4}$_{\tiny\begin{tcolorbox}[bluebox, left=0pt]{\text{$\downarrow \textbf{0.3}$}} \end{tcolorbox}}$ \\
\bottomrule
\end{tabular}%
}
\end{table*}

\section{Experiments}
\label{sec:experiments}

% --- TABLE 1: SUPERVISED LEARNING ---
\begin{table*}[t]
\centering
\caption{Supervised Learning results on OGBG-HIV and OGBG-PCBA.}
\label{tab:super}
\renewcommand\arraystretch{0.99}
\setlength\tabcolsep{6pt}
\resizebox{0.95\textwidth}{!}{
\begin{tabular}{l|cccc|cccc}
\toprule
\multirow{2}{*}{\textbf{Method}} & \multicolumn{4}{c|}{\textbf{OGBG-MOLHIV (ROC-AUC: 78.7)}} & \multicolumn{4}{c}{\textbf{OGBG-MOLPCBA (AP: 27.2)}} \\
\cline{2-9}
& 20\% & 30\% & 50\% & 70\% & 20\% & 30\% & 50\% & 70\% \\
\midrule
Random & 69.3$_{\tiny\begin{tcolorbox}[bluebox, left=0pt]{\text{$\downarrow \textbf{9.4}$}} \end{tcolorbox}}$ & 72.7$_{\tiny\begin{tcolorbox}[bluebox, left=0pt]{\text{$\downarrow \textbf{6.0}$}} \end{tcolorbox}}$ & 73.4$_{\tiny\begin{tcolorbox}[bluebox, left=0pt]{\text{$\downarrow \textbf{5.3}$}} \end{tcolorbox}}$ & 75.6$_{\tiny\begin{tcolorbox}[bluebox, left=0pt]{\text{$\downarrow \textbf{3.1}$}} \end{tcolorbox}}$ & 19.4$_{\tiny\begin{tcolorbox}[bluebox, left=0pt]{\text{$\downarrow \textbf{7.8}$}} \end{tcolorbox}}$ & 21.7$_{\tiny\begin{tcolorbox}[bluebox, left=0pt]{\text{$\downarrow \textbf{5.5}$}} \end{tcolorbox}}$ & 23.9$_{\tiny\begin{tcolorbox}[bluebox, left=0pt]{\text{$\downarrow \textbf{3.3}$}} \end{tcolorbox}}$ & 26.3$_{\tiny\begin{tcolorbox}[bluebox, left=0pt]{\text{$\downarrow \textbf{0.9}$}} \end{tcolorbox}}$ \\
CD & 72.6$_{\tiny\begin{tcolorbox}[bluebox, left=0pt]{\text{$\downarrow \textbf{6.1}$}} \end{tcolorbox}}$ & 73.0$_{\tiny\begin{tcolorbox}[bluebox, left=0pt]{\text{$\downarrow \textbf{5.7}$}} \end{tcolorbox}}$ & 75.3$_{\tiny\begin{tcolorbox}[bluebox, left=0pt]{\text{$\downarrow \textbf{3.4}$}} \end{tcolorbox}}$ & 76.7$_{\tiny\begin{tcolorbox}[bluebox, left=0pt]{\text{$\downarrow \textbf{2.0}$}} \end{tcolorbox}}$ & 18.0$_{\tiny\begin{tcolorbox}[bluebox, left=0pt]{\text{$\downarrow \textbf{9.2}$}} \end{tcolorbox}}$ & 20.7$_{\tiny\begin{tcolorbox}[bluebox, left=0pt]{\text{$\downarrow \textbf{6.5}$}} \end{tcolorbox}}$ & 21.7$_{\tiny\begin{tcolorbox}[bluebox, left=0pt]{\text{$\downarrow \textbf{5.5}$}} \end{tcolorbox}}$ & 26.4$_{\tiny\begin{tcolorbox}[bluebox, left=0pt]{\text{$\downarrow \textbf{1.0}$}} \end{tcolorbox}}$ \\
Herding & 69.5$_{\tiny\begin{tcolorbox}[bluebox, left=0pt]{\text{$\downarrow \textbf{9.2}$}} \end{tcolorbox}}$ & 73.3$_{\tiny\begin{tcolorbox}[bluebox, left=0pt]{\text{$\downarrow \textbf{5.4}$}} \end{tcolorbox}}$ & 74.5$_{\tiny\begin{tcolorbox}[bluebox, left=0pt]{\text{$\downarrow \textbf{4.2}$}} \end{tcolorbox}}$ & 75.9$_{\tiny\begin{tcolorbox}[bluebox, left=0pt]{\text{$\downarrow \textbf{2.8}$}} \end{tcolorbox}}$ & 13.3$_{\tiny\begin{tcolorbox}[bluebox, left=0pt]{\text{$\downarrow \textbf{13.9}$}} \end{tcolorbox}}$ & 14.0$_{\tiny\begin{tcolorbox}[bluebox, left=0pt]{\text{$\downarrow \textbf{13.2}$}} \end{tcolorbox}}$ & 17.8$_{\tiny\begin{tcolorbox}[bluebox, left=0pt]{\text{$\downarrow \textbf{9.4}$}} \end{tcolorbox}}$ & 23.0$_{\tiny\begin{tcolorbox}[bluebox, left=0pt]{\text{$\downarrow \textbf{4.2}$}} \end{tcolorbox}}$ \\
K-Center & 67.2$_{\tiny\begin{tcolorbox}[bluebox, left=0pt]{\text{$\downarrow \textbf{11.5}$}} \end{tcolorbox}}$ & 70.8$_{\tiny\begin{tcolorbox}[bluebox, left=0pt]{\text{$\downarrow \textbf{7.9}$}} \end{tcolorbox}}$ & 72.6$_{\tiny\begin{tcolorbox}[bluebox, left=0pt]{\text{$\downarrow \textbf{6.1}$}} \end{tcolorbox}}$ & 73.9$_{\tiny\begin{tcolorbox}[bluebox, left=0pt]{\text{$\downarrow \textbf{4.8}$}} \end{tcolorbox}}$ & 16.9$_{\tiny\begin{tcolorbox}[bluebox, left=0pt]{\text{$\downarrow \textbf{10.3}$}} \end{tcolorbox}}$ & 19.4$_{\tiny\begin{tcolorbox}[bluebox, left=0pt]{\text{$\downarrow \textbf{7.8}$}} \end{tcolorbox}}$ & 22.8$_{\tiny\begin{tcolorbox}[bluebox, left=0pt]{\text{$\downarrow \textbf{4.4}$}} \end{tcolorbox}}$ & 26.1$_{\tiny\begin{tcolorbox}[bluebox, left=0pt]{\text{$\downarrow \textbf{1.1}$}} \end{tcolorbox}}$ \\
SVP & 73.9$_{\tiny\begin{tcolorbox}[bluebox, left=0pt]{\text{$\downarrow \textbf{4.8}$}} \end{tcolorbox}}$ & 74.2$_{\tiny\begin{tcolorbox}[bluebox, left=0pt]{\text{$\downarrow \textbf{4.5}$}} \end{tcolorbox}}$ & 75.8$_{\tiny\begin{tcolorbox}[bluebox, left=0pt]{\text{$\downarrow \textbf{2.9}$}} \end{tcolorbox}}$ & 77.3$_{\tiny\begin{tcolorbox}[bluebox, left=0pt]{\text{$\downarrow \textbf{1.4}$}} \end{tcolorbox}}$ & 19.4$_{\tiny\begin{tcolorbox}[bluebox, left=0pt]{\text{$\downarrow \textbf{7.6}$}} \end{tcolorbox}}$ & 21.9$_{\tiny\begin{tcolorbox}[bluebox, left=0pt]{\text{$\downarrow \textbf{5.3}$}} \end{tcolorbox}}$ & 23.5$_{\tiny\begin{tcolorbox}[bluebox, left=0pt]{\text{$\downarrow \textbf{3.7}$}} \end{tcolorbox}}$ & 26.0$_{\tiny\begin{tcolorbox}[bluebox, left=0pt]{\text{$\downarrow \textbf{1.1}$}} \end{tcolorbox}}$ \\
Margin & 74.0$_{\tiny\begin{tcolorbox}[bluebox, left=0pt]{\text{$\downarrow \textbf{4.7}$}} \end{tcolorbox}}$ & 74.4$_{\tiny\begin{tcolorbox}[bluebox, left=0pt]{\text{$\downarrow \textbf{4.3}$}} \end{tcolorbox}}$ & 75.8$_{\tiny\begin{tcolorbox}[bluebox, left=0pt]{\text{$\downarrow \textbf{2.9}$}} \end{tcolorbox}}$ & 77.5$_{\tiny\begin{tcolorbox}[bluebox, left=0pt]{\text{$\downarrow \textbf{1.2}$}} \end{tcolorbox}}$ & 18.8$_{\tiny\begin{tcolorbox}[bluebox, left=0pt]{\text{$\downarrow \textbf{8.1}$}} \end{tcolorbox}}$ & 21.5$_{\tiny\begin{tcolorbox}[bluebox, left=0pt]{\text{$\downarrow \textbf{5.7}$}} \end{tcolorbox}}$ & 23.9$_{\tiny\begin{tcolorbox}[bluebox, left=0pt]{\text{$\downarrow \textbf{3.3}$}} \end{tcolorbox}}$ & 27.0$_{\tiny\begin{tcolorbox}[bluebox, left=0pt]{\text{$\downarrow \textbf{0.2}$}} \end{tcolorbox}}$ \\
Forgetting & 74.2$_{\tiny\begin{tcolorbox}[bluebox, left=0pt]{\text{$\downarrow \textbf{4.5}$}} \end{tcolorbox}}$ & 74.8$_{\tiny\begin{tcolorbox}[bluebox, left=0pt]{\text{$\downarrow \textbf{3.9}$}} \end{tcolorbox}}$ & 75.9$_{\tiny\begin{tcolorbox}[bluebox, left=0pt]{\text{$\downarrow \textbf{3.1}$}} \end{tcolorbox}}$ & 76.9$_{\tiny\begin{tcolorbox}[bluebox, left=0pt]{\text{$\downarrow \textbf{1.6}$}} \end{tcolorbox}}$ & 19.4$_{\tiny\begin{tcolorbox}[bluebox, left=0pt]{\text{$\downarrow \textbf{7.5}$}} \end{tcolorbox}}$ & 21.9$_{\tiny\begin{tcolorbox}[bluebox, left=0pt]{\text{$\downarrow \textbf{5.3}$}} \end{tcolorbox}}$ & 23.3$_{\tiny\begin{tcolorbox}[bluebox, left=0pt]{\text{$\downarrow \textbf{3.9}$}} \end{tcolorbox}}$ & 26.8$_{\tiny\begin{tcolorbox}[bluebox, left=0pt]{\text{$\downarrow \textbf{0.4}$}} \end{tcolorbox}}$ \\
GraNd-4 & 73.8$_{\tiny\begin{tcolorbox}[bluebox, left=0pt]{\text{$\downarrow \textbf{4.9}$}} \end{tcolorbox}}$ & 74.2$_{\tiny\begin{tcolorbox}[bluebox, left=0pt]{\text{$\downarrow \textbf{4.5}$}} \end{tcolorbox}}$ & 75.8$_{\tiny\begin{tcolorbox}[bluebox, left=0pt]{\text{$\downarrow \textbf{3.4}$}} \end{tcolorbox}}$ & 77.5$_{\tiny\begin{tcolorbox}[bluebox, left=0pt]{\text{$\downarrow \textbf{1.2}$}} \end{tcolorbox}}$ & 18.0$_{\tiny\begin{tcolorbox}[bluebox, left=0pt]{\text{$\downarrow \textbf{9.2}$}} \end{tcolorbox}}$ & 21.2$_{\tiny\begin{tcolorbox}[bluebox, left=0pt]{\text{$\downarrow \textbf{6.3}$}} \end{tcolorbox}}$ & 23.6$_{\tiny\begin{tcolorbox}[bluebox, left=0pt]{\text{$\downarrow \textbf{3.6}$}} \end{tcolorbox}}$ & 26.9$_{\tiny\begin{tcolorbox}[bluebox, left=0pt]{\text{$\downarrow \textbf{0.3}$}} \end{tcolorbox}}$ \\
DeepFool & 72.2$_{\tiny\begin{tcolorbox}[bluebox, left=0pt]{\text{$\downarrow \textbf{6.5}$}} \end{tcolorbox}}$ & 73.3$_{\tiny\begin{tcolorbox}[bluebox, left=0pt]{\text{$\downarrow \textbf{5.4}$}} \end{tcolorbox}}$ & 74.9$_{\tiny\begin{tcolorbox}[bluebox, left=0pt]{\text{$\downarrow \textbf{3.4}$}} \end{tcolorbox}}$ & 75.5$_{\tiny\begin{tcolorbox}[bluebox, left=0pt]{\text{$\downarrow \textbf{3.2}$}} \end{tcolorbox}}$ & 17.6$_{\tiny\begin{tcolorbox}[bluebox, left=0pt]{\text{$\downarrow \textbf{9.6}$}} \end{tcolorbox}}$ & 21.3$_{\tiny\begin{tcolorbox}[bluebox, left=0pt]{\text{$\downarrow \textbf{6.2}$}} \end{tcolorbox}}$ & 23.2$_{\tiny\begin{tcolorbox}[bluebox, left=0pt]{\text{$\downarrow \textbf{4.0}$}} \end{tcolorbox}}$ & 26.5$_{\tiny\begin{tcolorbox}[bluebox, left=0pt]{\text{$\downarrow \textbf{1.0}$}} \end{tcolorbox}}$ \\
CRAIG & 73.5$_{\tiny\begin{tcolorbox}[bluebox, left=0pt]{\text{$\downarrow \textbf{5.0}$}} \end{tcolorbox}}$ & 74.4$_{\tiny\begin{tcolorbox}[bluebox, left=0pt]{\text{$\downarrow \textbf{4.3}$}} \end{tcolorbox}}$ & 75.9$_{\tiny\begin{tcolorbox}[bluebox, left=0pt]{\text{$\downarrow \textbf{2.7}$}} \end{tcolorbox}}$ & 77.5$_{\tiny\begin{tcolorbox}[bluebox, left=0pt]{\text{$\downarrow \textbf{1.2}$}} \end{tcolorbox}}$ & \textbf{22.5}$_{\tiny\begin{tcolorbox}[bluebox, left=0pt]{\text{$\downarrow \textbf{4.5}$}} \end{tcolorbox}}$ & 24.5$_{\tiny\begin{tcolorbox}[bluebox, left=0pt]{\text{$\downarrow \textbf{2.5}$}} \end{tcolorbox}}$ & 24.7$_{\tiny\begin{tcolorbox}[bluebox, left=0pt]{\text{$\downarrow \textbf{2.2}$}} \end{tcolorbox}}$ & 27.1$_{\tiny\begin{tcolorbox}[bluebox, left=0pt]{\text{$\downarrow \textbf{1.0}$}} \end{tcolorbox}}$ \\
GLSTER & 73.6$_{\tiny\begin{tcolorbox}[bluebox, left=0pt]{\text{$\downarrow \textbf{4.9}$}} \end{tcolorbox}}$ & 75.0$_{\tiny\begin{tcolorbox}[bluebox, left=0pt]{\text{$\downarrow \textbf{3.7}$}} \end{tcolorbox}}$ & 75.8$_{\tiny\begin{tcolorbox}[bluebox, left=0pt]{\text{$\downarrow \textbf{2.8}$}} \end{tcolorbox}}$ & 77.0$_{\tiny\begin{tcolorbox}[bluebox, left=0pt]{\text{$\downarrow \textbf{0.7}$}} \end{tcolorbox}}$ & \textbf{22.5}$_{\tiny\begin{tcolorbox}[bluebox, left=0pt]{\text{$\downarrow \textbf{4.5}$}} \end{tcolorbox}}$ & \textbf{24.8}$_{\tiny\begin{tcolorbox}[bluebox, left=0pt]{\text{$\downarrow \textbf{2.2}$}} \end{tcolorbox}}$ & 24.2$_{\tiny\begin{tcolorbox}[bluebox, left=0pt]{\text{$\downarrow \textbf{2.7}$}} \end{tcolorbox}}$ & 27.0$_{\tiny\begin{tcolorbox}[bluebox, left=0pt]{\text{$\downarrow \textbf{0.2}$}} \end{tcolorbox}}$ \\
Influence & 72.9$_{\tiny\begin{tcolorbox}[bluebox, left=0pt]{\text{$\downarrow \textbf{5.6}$}} \end{tcolorbox}}$ & 73.7$_{\tiny\begin{tcolorbox}[bluebox, left=0pt]{\text{$\downarrow \textbf{5.0}$}} \end{tcolorbox}}$ & 74.8$_{\tiny\begin{tcolorbox}[bluebox, left=0pt]{\text{$\downarrow \textbf{3.9}$}} \end{tcolorbox}}$ & 77.4$_{\tiny\begin{tcolorbox}[bluebox, left=0pt]{\text{$\downarrow \textbf{1.3}$}} \end{tcolorbox}}$ & 17.7$_{\tiny\begin{tcolorbox}[bluebox, left=0pt]{\text{$\downarrow \textbf{9.3}$}} \end{tcolorbox}}$ & 23.5$_{\tiny\begin{tcolorbox}[bluebox, left=0pt]{\text{$\downarrow \textbf{3.7}$}} \end{tcolorbox}}$ & 23.5$_{\tiny\begin{tcolorbox}[bluebox, left=0pt]{\text{$\downarrow \textbf{3.7}$}} \end{tcolorbox}}$ & 26.6$_{\tiny\begin{tcolorbox}[bluebox, left=0pt]{\text{$\downarrow \textbf{0.6}$}} \end{tcolorbox}}$ \\
EL2N-20 & 74.0$_{\tiny\begin{tcolorbox}[bluebox, left=0pt]{\text{$\downarrow \textbf{4.7}$}} \end{tcolorbox}}$ & 75.5$_{\tiny\begin{tcolorbox}[bluebox, left=0pt]{\text{$\downarrow \textbf{3.2}$}} \end{tcolorbox}}$ & 76.9$_{\tiny\begin{tcolorbox}[bluebox, left=0pt]{\text{$\downarrow \textbf{1.9}$}} \end{tcolorbox}}$ & 77.7$_{\tiny\begin{tcolorbox}[bluebox, left=0pt]{\text{$\downarrow \textbf{1.1}$}} \end{tcolorbox}}$ & 19.1$_{\tiny\begin{tcolorbox}[bluebox, left=0pt]{\text{$\downarrow \textbf{8.1}$}} \end{tcolorbox}}$ & 22.9$_{\tiny\begin{tcolorbox}[bluebox, left=0pt]{\text{$\downarrow \textbf{3.9}$}} \end{tcolorbox}}$ & 25.0$_{\tiny\begin{tcolorbox}[bluebox, left=0pt]{\text{$\downarrow \textbf{2.0}$}} \end{tcolorbox}}$ & 26.6$_{\tiny\begin{tcolorbox}[bluebox, left=0pt]{\text{$\downarrow \textbf{1.1}$}} \end{tcolorbox}}$ \\
DP & 72.0$_{\tiny\begin{tcolorbox}[bluebox, left=0pt]{\text{$\downarrow \textbf{6.7}$}} \end{tcolorbox}}$ & 74.1$_{\tiny\begin{tcolorbox}[bluebox, left=0pt]{\text{$\downarrow \textbf{4.6}$}} \end{tcolorbox}}$ & 76.0$_{\tiny\begin{tcolorbox}[bluebox, left=0pt]{\text{$\downarrow \textbf{2.7}$}} \end{tcolorbox}}$ & 76.9$_{\tiny\begin{tcolorbox}[bluebox, left=0pt]{\text{$\downarrow \textbf{1.8}$}} \end{tcolorbox}}$ & 19.6$_{\tiny\begin{tcolorbox}[bluebox, left=0pt]{\text{$\downarrow \textbf{7.6}$}} \end{tcolorbox}}$ & 21.5$_{\tiny\begin{tcolorbox}[bluebox, left=0pt]{\text{$\downarrow \textbf{5.7}$}} \end{tcolorbox}}$ & 24.9$_{\tiny\begin{tcolorbox}[bluebox, left=0pt]{\text{$\downarrow \textbf{2.3}$}} \end{tcolorbox}}$ & 26.4$_{\tiny\begin{tcolorbox}[bluebox, left=0pt]{\text{$\downarrow \textbf{0.8}$}} \end{tcolorbox}}$ \\
\midrule
\textbf{GraphSculptor} & \textbf{74.6}$_{\tiny\begin{tcolorbox}[bluebox, left=0pt]{\text{$\downarrow \textbf{4.1}$}} \end{tcolorbox}}$ & \textbf{75.6}$_{\tiny\begin{tcolorbox}[bluebox, left=0pt]{\text{$\downarrow \textbf{3.1}$}} \end{tcolorbox}}$ & \textbf{77.0}$_{\tiny\begin{tcolorbox}[bluebox, left=0pt]{\text{$\downarrow \textbf{1.7}$}} \end{tcolorbox}}$ & \textbf{78.9}$_{\tiny\begin{tcolorbox}[pinkbox, left=0pt]{\text{$\uparrow \textbf{0.2}$}} \end{tcolorbox}}$ & 19.5$_{\tiny\begin{tcolorbox}[bluebox, left=0pt]{\text{$\downarrow \textbf{7.7}$}} \end{tcolorbox}}$ & 22.4$_{\tiny\begin{tcolorbox}[bluebox, left=0pt]{\text{$\downarrow \textbf{4.8}$}} \end{tcolorbox}}$ & \textbf{25.3}$_{\tiny\begin{tcolorbox}[bluebox, left=0pt]{\text{$\downarrow \textbf{1.9}$}} \end{tcolorbox}}$ & \textbf{27.4}$_{\tiny\begin{tcolorbox}[pinkbox, left=0pt]{\text{$\uparrow \textbf{0.2}$}} \end{tcolorbox}}$ \\
\bottomrule
\end{tabular}
}
\end{table*}

\begin{table}[t]
\centering
\caption{Supervised Learning results on COLLAB and IMDB-BINARY.}
\label{tab:super}
\renewcommand\arraystretch{1.25}
\setlength\tabcolsep{4pt}
\resizebox{0.49\textwidth}{!}{
\begin{tabular}{l|cccc|cccc}
\toprule
\multirow{2}{*}{\textbf{Method}} & \multicolumn{4}{c|}{\textbf{COLLAB (81.4)}} & \multicolumn{4}{c}{\textbf{IMDB-BINARY (77.4)}} \\
& 20\% & 30\% & 50\% & 70\% & 20\% & 30\% & 50\% & 70\% \\
\midrule
Random & 75.1 & 76.2 & 80.4 & 80.9 & 73.8 & 74.2 & 74.6 & 75.4 \\
CD & 74.5 & 76.4 & 80.5 & 80.7 & 73.5 & 74.2 & 75.0 & 75.5 \\
Herding & 75.6 & 76.5 & 80.7 & 80.2 & 74.0 & 74.5 & 75.2 & 76.0 \\
K-Center & 75.4 & 75.8 & 78.6 & 79.8 & 74.2 & 74.7 & \textbf{75.5} & \textbf{76.1} \\
\midrule
\textbf{GraphSculptor} & \textbf{76.0} & \textbf{76.6} & \textbf{81.0} & \textbf{81.3} & \textbf{74.2} & \textbf{75.0} & 75.3 & 75.8 \\
\bottomrule
\end{tabular}
}
\end{table}

\subsection{Experimental Setup}
\label{sec:setup}

\paragraph{Datasets and Tasks.}
Our evaluation covers two learning paradigms.
\textbf{Pre-training and Transfer Learning.} For graph SSL, we pre-train on ZINC, which contains $\sim$250{,}000 commercially available molecules~\cite{zinc15}. We then transfer the pre-trained models to eight binary classification tasks from MoleculeNet~\cite{moleculenet}. For all MoleculeNet tasks, we use the standard scaffold splits to evaluate generalization.
\textbf{Supervised Learning.} To assess the generality of our selection strategy beyond SSL pre-training, we additionally conduct supervised graph classification on large-scale OGB molecular benchmarks, OGBG-MOLHIV and OGBG-MOLPCBA~\cite{ogb-dataset}, following the official data splits provided by OGB. We further include two social graph benchmarks from TUDataset\cite{tu-dataset}, COLLAB and IMDB-BINARY, using the standard evaluation protocol. We benchmark GraphSculptor against 13 static data selection baselines, covering Random sampling, classical coreset methods (k-Center, Herding, CD), and representative label-informed pruning approaches (Influence, DP). 
. 
\paragraph{Implementation Details.}
\label{par:imp_details}
We use standard GNN frameworks for all experiments. For SSL pre-training on ZINC, we employ GraphMAE~\cite{graphmae} and evaluate via linear probing on MoleculeNet. For supervised learning on OGB, we train GraphGPS~\cite{graphgps} models from scratch on selected coresets. raphSculptor performs graph-to-text translation and semantic embedding in an offline manner. For molecular graphs, we use BioT5-base~\cite{biot5} for graph-to-text translation and SciBERT~\cite{scibert} as the frozen text encoder. For social-network datasets from TUDataset (COLLAB and IMDB-BINARY), we use a general-purpose LLM (\textit{e.g.}, Gemini) for graph-to-text translation (using fixed prompts reported in the Appendix) and a BERT as the encoder. The number of clusters $K$ is treated as a hyperparameter; the tuning protocol and selected values are reported in the Appendix. All reported results are the mean and standard deviation over 5 random seeds.

\subsection{Main Result in Graph SSL}
\label{sec:res_ssl}
We evaluate whether a curated subset can match full-data pre-training for graph SSL. Table~\ref{tab:trans} reports transfer results on eight MoleculeNet benchmarks (ROC-AUC). GraphSculptor remains competitive across budgets and achieves the highest average ROC-AUC in Table~\ref{tab:trans} under each reported ratio: 74.1 at 70\% remaining (vs.\ 73.7 for full-data pre-training), 73.8 at 50\% remaining, and 73.4 at 10\% remaining (vs.\ 72.3--72.6 for classical coreset baselines such as Herding, K-Center, and CD). Two observations stand out. First, GraphSculptor matches or slightly exceeds full-data pre-training at moderate budgets (\textit{e.g.}, 74.1 vs.\ 73.7 at 70\%), suggesting that removing redundant graphs may sharpen the effective pre-training signal. Second, its advantage becomes more pronounced under aggressive compression (\textit{e.g.}, 73.4 at 10\% vs.\ 72.3--72.6 for classical baselines), consistent with our theory that the loss-gap bound scales with $1/M_{\text{target}}$ and depends on the allocation across clusters, and with our design that uses fused structural--semantic clustering and cluster-aware budget allocation to better cover diverse, non-redundant regions.

\subsection{Generality in Supervised Learning}
\label{sec:res_supervised}

We further evaluate the generality of GraphSculptor on supervised graph classification. On OGBG-MOLHIV and OGBG-MOLPCBA, GraphSculptor performs label-free selection while several strong baselines leverage labels for pruning; nevertheless, GraphSculptor remains competitive and often yields better accuracy under the same budget. For example, on MOLHIV it matches or slightly exceeds the full-data result at the 70\% budget, and at 50\% it achieves 77.0 on HIV and 25.3 on PCBA, outperforming Random subsets even at larger retention ratios. Beyond molecular graphs, we also test social-network benchmarks from TUDataset (Table~\ref{tab:super}). On COLLAB, GraphSculptor consistently improves over Random and classical coreset baselines across budgets, reaching 81.0/81.3 at 50\%/70\% and closely tracking the full-data reference (81.4). On IMDB-BINARY, the gains are more modest, which is expected given the small dataset size and limited redundancy. Nonetheless, GraphSculptor is most effective in the low-budget regime, outperforming Random and achieving the best (or tied-best) accuracy at 20\%--30\% retention (75.0 at 30\%), while at higher budgets (50\%--70\%) it becomes comparable to classical coreset methods such as K-Center and Herding.
Overall, these results suggest that diversity-driven, label-free selection can serve as a broadly applicable preprocessing step for reducing training cost while maintaining supervised performance across graph domains.

\subsection{Further Discussion}

\begin{figure}[!t] % !ht% \setlength{\abovecaptionskip}{-0.1cm}   %调整图片标题与图距离
\begin{center}
\includegraphics[width=0.99\linewidth]{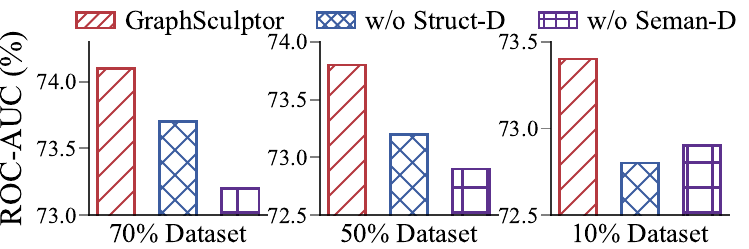}
\end{center}
\caption{\textbf{Ablation study} of GraphSculptor. ``w/o Struct-D'' denotes the absence of structural diversity; ``w/o Seman-D'' denotes the lack of semantic diversity.}
\label{fig:ablation}
\end{figure}

\paragraph{Ablation Study.} We study the contribution of the two diversity views in GraphSculptor by comparing the full method with two ablations: Structure-only, which performs clustering and selection using only structural descriptors $\mathbf{z}_{S}$, and Semantics-only, which uses only semantic embeddings $\mathbf{z}_{T}$. As shown in Figure~\ref{fig:ablation}, the full model achieves the best performance across all budgets, while removing either view leads to a consistent degradation, indicating that the two views provide complementary signals for coreset curation. In our setting, dropping the semantic view typically incurs a larger performance drop than dropping the structural view, suggesting that language-based semantics can capture information not well reflected by structural statistics alone. Overall, combining both structural and semantic cues yields a more reliable criterion than either single view.

\begin{figure}[!t] % !ht% \setlength{\abovecaptionskip}{-0.1cm}   %调整图片标题与图距离
\begin{center}
\includegraphics[width=0.99\linewidth]{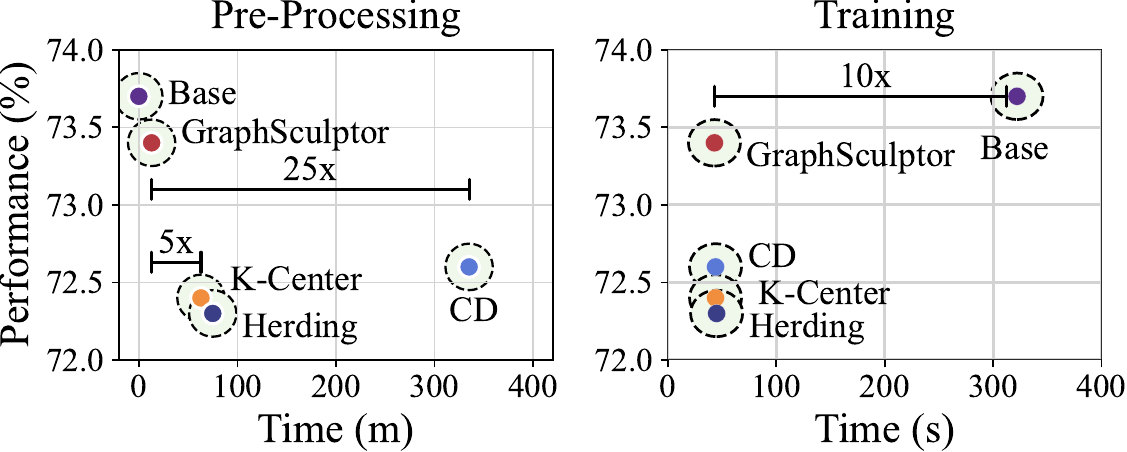}
\end{center}
\caption{\textbf{Efficiency analysis}. }
\label{fig:effic}
\end{figure}

\paragraph{Efficiency Analysis.} Figure~\ref{fig:effic} summarizes the end-to-end efficiency of GraphSculptor. The method incurs a one-time preprocessing cost for graph-to-text and text encoding (about 30 minutes in our setup), but this cost is amortized across pre-training runs and is substantially lower than several training-signal-based pruning baselines (5--25$\times$ faster in wall-clock time). More importantly, coreset pre-training provides a direct reduction in training time proportional to the retained data budget: for example, using a 10\% coreset cuts pre-training time by nearly 90\% compared to full-data pre-training, while maintaining comparable downstream performance (and occasionally yielding small gains). Overall, GraphSculptor offers a favorable performance--efficiency trade-off by shifting selection to an offline, model-agnostic step and reducing the dominant cost of full-data pre-training.

\section{Conclusion}
We revisit the prevailing practice of scaling unlabeled graph corpora for self-supervised pre-training and observe notable redundancy in our benchmark evaluation. We propose GraphSculptor, a label-free and model-agnostic framework for constructing compact pre-training coresets by combining a structural view based on intrinsic graph statistics with a complementary semantic view derived from graph-to-text and a frozen text encoder. We further provide a loss-gap bound under a standard linearized optimization regime, offering theoretical motivation for cluster-aware budget allocation. Empirically, GraphSculptor achieves favorable performance--efficiency trade-offs across molecular and social graph benchmarks. Overall, our results suggest that diversity-driven, offline data curation can serve as a practical preprocessing step for more efficient scaling of graph SSL.  As future work, we will explore more reliable semantic abstractions for text-scarce graphs and strengthen the theoretical analysis in training dynamics.

\bibliographystyle{named}
\bibliography{ijcai26}

\end{document}